\def\year{2022}\relax
\documentclass[letterpaper]{article} % DO NOT CHANGE THIS
\usepackage{arxiv}  % DO NOT CHANGE THIS
\usepackage{times}  % DO NOT CHANGE THIS
\usepackage{helvet}  % DO NOT CHANGE THIS
\usepackage{courier}  % DO NOT CHANGE THIS
\usepackage[hyphens]{url}  % DO NOT CHANGE THIS
\usepackage{graphicx} % DO NOT CHANGE THIS
\usepackage{natbib}  % DO NOT CHANGE THIS AND DO NOT ADD ANY OPTIONS TO IT
\usepackage{caption} % DO NOT CHANGE THIS AND DO NOT ADD ANY OPTIONS TO IT
\DeclareCaptionStyle{ruled}{labelfont=normalfont,labelsep=colon,strut=off} % DO NOT CHANGE THIS
\usepackage{amsthm}

\newcommand{\tamp}{{\sc tamp}}
\newcommand{\gentamp}{{\sc GenTAMP}}
\newcommand{\pddlstream}{{\sc PDDLStream}}

\newcommand{\pre}{\mathit{pre}}
\newcommand{\eff}{\mathit{eff}}

\usepackage{amsmath}
\usepackage{amssymb}
\usepackage{algorithm2e}
\usepackage{listings}

\newtheorem{definition}{Definition}

\usepackage{todonotes}

\title{Discovering State and Action Abstractions for \\ Generalized Task and Motion Planning}
\author {
    Aidan Curtis,
    Tom Silver,
    Joshua B. Tenenbaum,
    Tom\'{a}s Lozano-P\'{e}rez,
    Leslie Pack Kaelbling
}
\affiliations {
    MIT Computer Science and Artificial Intelligence Laboratory
}
\usepackage{bibentry}
\begin{document}

\maketitle

\begin{abstract}
Generalized planning accelerates classical planning by finding an algorithm-like policy that solves multiple instances of a task. A generalized plan can be learned from a few training examples and applied to an entire domain of problems. Generalized planning approaches perform well in discrete AI planning problems that involve large numbers of objects and extended action sequences to achieve the goal. In this paper, we propose an algorithm for learning features, abstractions, and generalized plans for continuous robotic task and motion planning (TAMP) and examine the unique difficulties that arise when forced to consider geometric and physical constraints as a part of the generalized plan. Additionally, we show that these simple generalized plans learned from only a handful of examples can be used to improve the search efficiency of \tamp{} solvers. 
\end{abstract}

\section{Introduction}
\label{sec:introduction}

% Highest level introduction which gives motivation for the paper
A shared goal in the robotics and artificial intelligence communities is to build intelligent autonomous robotic agents that can solve long-horizon tasks in arbitrary human environments. Two of the major roadblocks to building such general-purpose intelligent systems are {\em scalability} to settings with large numbers of objects and {\em generalizability} to a wide array of novel environments and goals.

A scalable and generalizable intelligent robotic agent must consider objects' geometric, physical, visual, and semantic attributes to select and execute goal-directed actions. For example, a task such as ``pack all my camping gear into my backpack'' might require reasoning about the location, size and shape, mass and deformability, ownership, and category designation of each item in the environment. The agent must also execute actions that progress to the specified goal without violating physical and geometric constraints induced by the environment.

One typical approach to solving this type of problem is task and motion planning (\tamp{})~\cite{garrett2020integrated}. \tamp{} solvers work by integrating discrete symbolic planning with geometric motion planning where the results from each of these two subsystems can affect the other's feasibility. Because of this mutual dependence between the symbolic and geometric constraints, \tamp{} solvers usually rely on iterative procedures that alternate between geometric and discrete symbolic planning until a geometrically feasible plan is found that reaches the goal. The resulting iterative planning algorithms are much more computationally complex than classical planning algorithms. While \tamp{} solvers are fully generalizable to novel environments and tasks, they are limited in scalability due to the large combinatorial and continuous search problem that exists for each problem.
\begin{figure}
\centering
    \includegraphics[width=\linewidth]{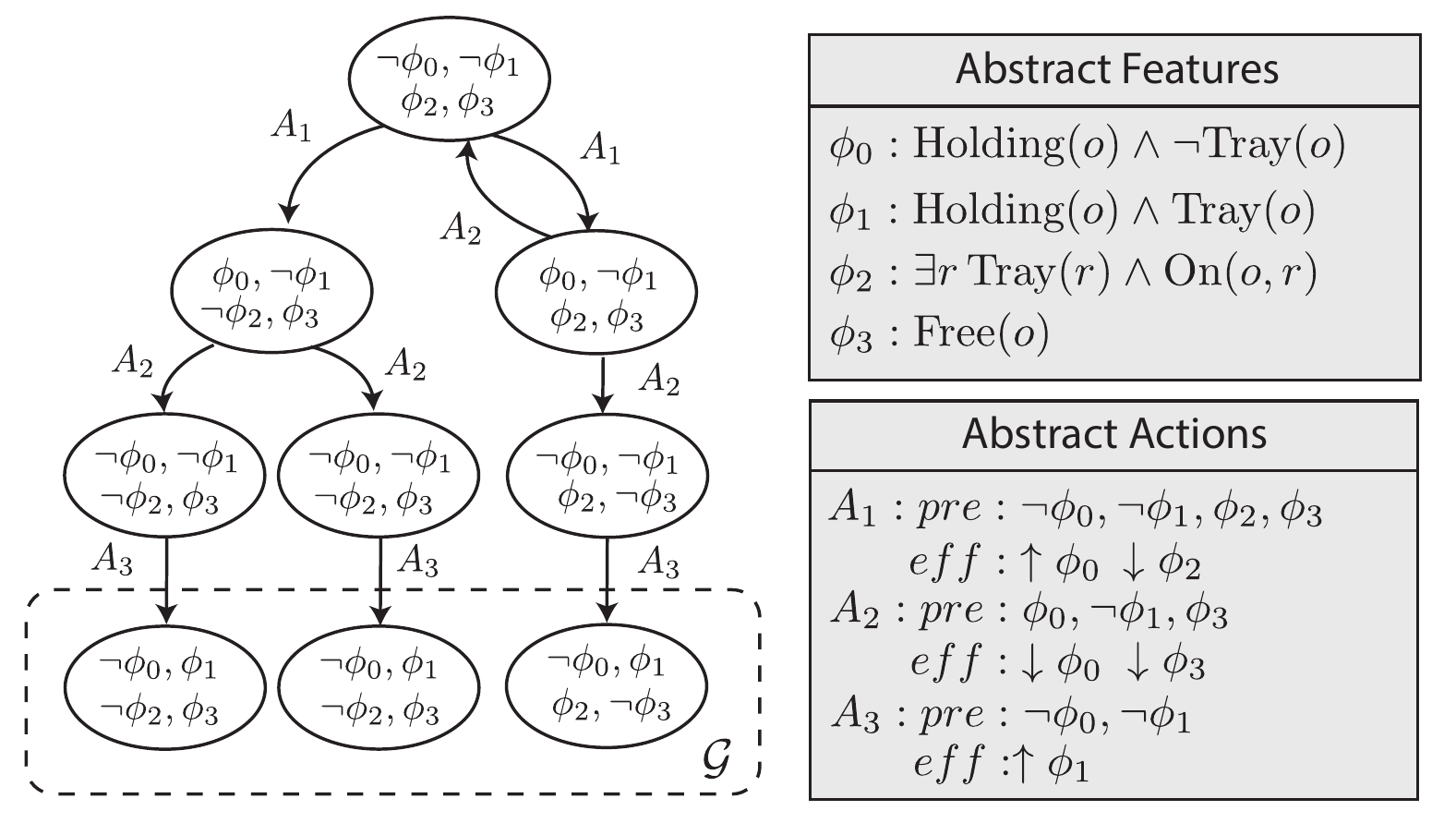}
    \caption{An abstract policy for the working \textsc{Load} task example. The generalized plan that solves this \textsc{Load} task picks an object ($A_1$) and places it on the tray ($A_2$) until either the tray is full ($\lnot\phi_3$) or no objects are remaining on the table ($\lnot\phi_2$). After one of these two features evaluates to false, the tray is grasped ($A_3$). Here $\texttt{Free}(o)$ is used as a shorthand for the larger feature $\exists{p, r}\forall{o_2, p_2}:\texttt{Tray}(r) \land \lnot \texttt{On}(o, p, r) \land \texttt{On}(o_2, p_2, r) \Rightarrow \lnot \texttt{CFree}(o, p, o_2, p_2, r)$
    % \tsnote{just something to consider: you could move this figure down to where you introduce the Load example, since it's not referenced in the intro. but if you leave it here that's fine too}
    % \arcnote{I think I heard somewhere that it's always good to have a figure on the first page}
    }
    \label{fig:example}
\end{figure}
\begin{figure*}
\centering
    \includegraphics[width=\linewidth]{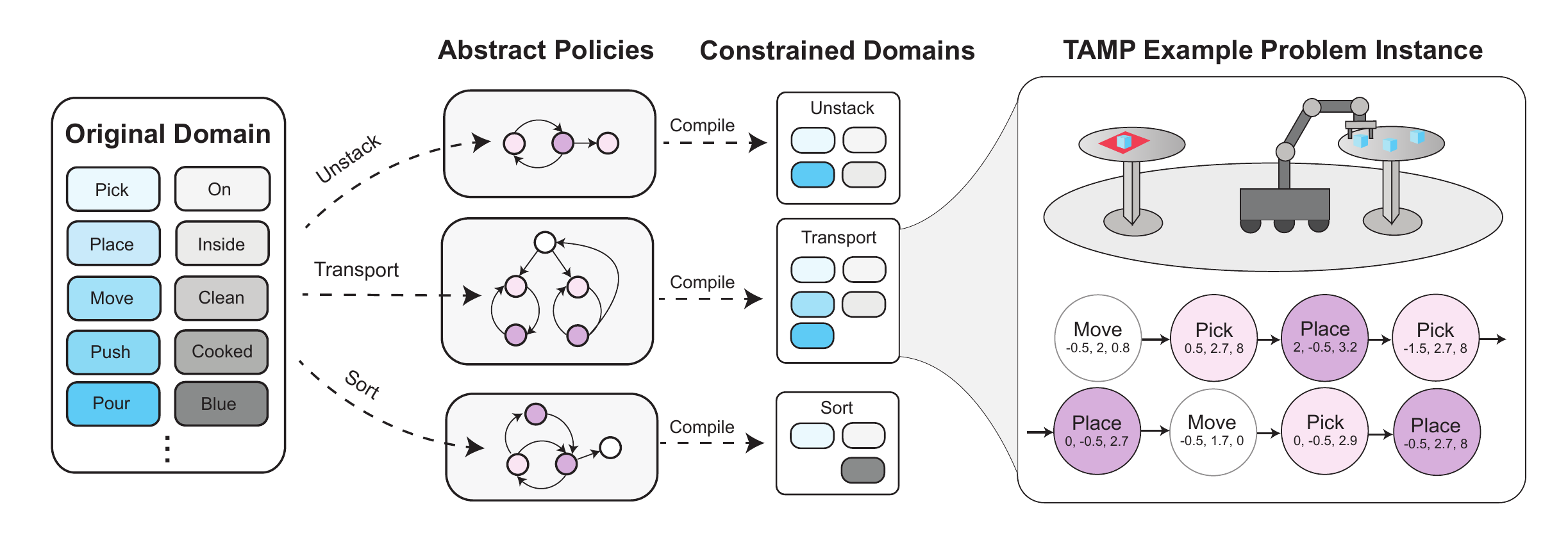}
    \caption{\textsc{GenTAMP} conceptual overview. Several robotics tasks are defined within a single domain. For each task, a policy is learned from a small set of example plans and compiled into a set of action and ordering constraints added to the original domain to create a task-specific domain. This task-specific domain forces \tamp{} search to follow the abstract policy, thus reducing the search space leading to faster planning.}
    \label{fig:concept}
\end{figure*}

% Generalized planning is great (srivastava)
Generalized planning is a method for tackling exactly this scalability problem in classical planning. 
Generalized planners find abstract high-level plans that apply to a large, often infinite, class of problem instances with a shared structure. For example, ``While there are objects in the box, pick up a reachable object in the box and place it on the table.'' is a generalized plan that applies to settings with any number of objects.
Solving each problem instance independently with standard planning approaches can be highly inefficient as planners are often slow and become exponentially slower with increasing solution length. 
The motivation of generalized planning is to avoid search in large state-space problem instances by identifying and exploiting structure in a few example problems with smaller state spaces. 
In addition, because languages for expressing generalized plans often include looping constructs, the resulting generalized plans are often more compact than traditional plans for large domains. While discrete generalized planning methods help in problems with large numbers of objects, they do not incorporate continuous features. 
As such, they would not be capable of discovering generalized plans that rely on continuous features of the environment, such as the remaining capacity of a backpack, the water level of a camping thermos, or more subtle collision constraints present in any robotic setting. While there are various approaches to generalized planning for classical AI planning domains, our method is the first to apply these approaches to continuous robotics domains in which the generalized plan requires complex continuous features. 

We propose \textsc{GenTAMP}, a framework for learning and executing generalized plans in mixed discrete-continuous environments. The novel contributions of this paper are as follows. First, we formulate the problem of generalized task and motion planning. Second, we outline a new algorithm for learning a generalized plan from examples that requires evaluation of complex continuous features with high-arity predicates and nested quantification. Third, we introduce an approach for executing generalized plans on unseen task and motion planning problems. Lastly, we evaluate our algorithm's ability to scale to large unseen problems and compare planning speeds with a standard \tamp{} solver.

\section{Background}
\label{sec:problem_formulation}

\textbf{Classical AI planning}
\label{sec:classplan}
A classical planning problem is typically defined by the tuple $P=\langle{\mathcal{S},I,A,G\rangle}$, where $\mathcal{S}$ is a set of state variables that can take a finite set of values, $I$ is the initial state, an initial assignment of values to each of the state variables, $A$ is a set of actions, and $G$ is a subset of $\mathcal{S}$ describing the goal state. A classical planning domain is defined by a deterministic successor state function $f: {\mathcal{S}}\times A \rightarrow{\mathcal{S}}$. The objective of classical planning is to find a sequence of actions and their resulting states $(a_1, s_1), (a_2, s_2), \dots, (a_n, s_n)$ such that $a_i\in{A}$, $s_{i+1} = f(s_i, a_{i+1})$, and $s_n\in{G}$. 

Such classical planning problems typically have an associated {\em lifted} task representation, which allows the domain description to abstract away from the particular set of entities in any problem instance. This description is parameterized by an entity set $\mathcal{O}$, and consists of a number of N-ary predicates $p(\bar{o}):\mathcal{O}^N\rightarrow\{0, 1\}$ and a set of action schemas, $A_i(\bar{o})$ with entity-parameterized positive and negative predicate preconditions ($\pre^+, \pre^-$) and effects ($\eff^+, \eff^-$) such that action $A_i(\bar{o})$ is applicable in state $s$ if its positive preconditions are satisfied in $s$ and its negative ones are not: %$\exists{\bar{o}\subseteq{\mathcal{O}}}\;
$\pre^+(A_i(\bar{o}))\subseteq{s}\land \pre^-(A_i(\bar{o}))\cap{s}=\emptyset$.  The resulting effect of applying the entity-parameterized action schema on a state $s$ removes the negative effects and adds the positive ones:  $f(s,A_i(\bar{o}))=(s\setminus \eff^-(A_i(\bar{o})))\cup \eff^+(A_i(\bar{o}))$. 
Given a particular set of entities $\mathcal{O}$ for a problem instance, before planning, this lifted representation is usually compiled down to the standard state-variable representation above in a process called {\em grounding}, in which each state variable corresponds to a {\em ground literal}, which is an application of a predicate to a tuple of entities. 

\noindent
\textbf{Generalized planning.}
\label{sec:genplan}
The generalized planning problem builds on this formulation and establishes a meta-problem $Q = \langle P_1, P_2, \dots \rangle$ encompassing a (possibly infinite) set of classical planning problems. The objective of generalized planning is to find a deterministic policy $\pi:\mathcal{S}\to{A}$ that, when successively applied starting from an initial state, results in a feasible action sequence that eventually reaches the goal for as many $P_i\in{Q}$ as possible. We model our problem formulation on the one established by~\citet{genplan_representation}, in which problem instances have no shared state or action representation $P_i = \langle{\mathcal{S}_i,I_i,A_i,G_i}\rangle$.

\noindent
\textbf{Task and motion planning.}
\label{sec:tamp}
The generalized planning methods in this paper build on the ability to solve \tamp{} problems; we use the \pddlstream{} ~\cite{pddlstream} problem definition language and associated algorithms as the basis for generalized planning in our implementations, but the overall approach is compatible with other \tamp{} methods. This \tamp{} problem definition amends the entity set to include \textit{sampled entities} such as grasps, collision-free pose trajectories, and object placement samples. These entities are sampled from high-dimensional nonlinear continuous spaces by \textit{streams}. Streams are continuous samplers with both procedural and declarative components. The procedural component is a function $g(\bar{o})$ that takes in a tuple of objects and generates sampled entities from some continuous space conditioned on the inputs. The declarative component specifies the semantics of these sampled entities and how they relate to existing entities in the domain through \textit{certified predicates} which have, as arguments, the entities sampled by the stream and are only referenced in initial states, goals, and the preconditions of actions. 
% Below is an example of the declarative component of a stream that takes in a surface and an object and generates valid continuous-valued placements of the object on that surface.
% \begin{small} % footnotesize
% \begin{lstlisting}
% SamplePlacement
%  |{\bf inputs}:| ($o, s$)
%  |{\bf domain}:| Movable($o$), Surface($s$)
%  |{\bf outputs}:| ($p$)
%  |{\bf cert}:| Pose($o, p, s$)
%  \end{lstlisting}
% \end{small}
Algorithms that use the \pddlstream{} problem definition typically start with an insufficient set of sampled entities and certified predicates. During planning, they sample additional entities, adding corresponding certified predicates until the goal is achievable by a classical planner.
%Because of the continuous nature of these environments, there are infinitely many possible sampled entities. As a result, we can no longer assume a closed world, and the non-existence of a sampled entity does not imply its negation. This point is of particular importance when evaluating features involving quantification. Algorithms using the \pddlstream{} formulation are still probabilistically semi-complete assuming that the lower-level samplers and planners it builds on are, themselves, probabilistically complete ~\cite{pddlstream}.

\noindent
\subsection{Generalized Task and Motion Planning}
In this work, we propose the generalized \tamp{} problem setting. Typical robotics applications that rely on \tamp{} algorithms have a single \textit{domain} that details the preconditions and effects of every action available to the agent and every possible property of objects and their relationships but does not specify the universe of objects, a concrete initial state, or a goal. Generalized \tamp{} additionally specifies a set of \textit{tasks} $\{Q_1,\dots,Q_N\}$ where each $Q$ is associated with a tuple $\langle{I_Q},{G_Q}\rangle$. $I_Q$ is a set of possible initial states in $Q$ (e.g. ``all red objects are on the table''), and $G_Q$ is a set of possible goals for $Q$ (e.g. ``all red objects are on the red mat''). Because these are infinite sets, they are described by first-order logical expressions that quantify over objects in the state and can therefore be applied to states containing arbitrary numbers of entities. Infinitely many \textit{problems} can be derived from any task with each problem instance $P_i$ containing a unique ground initial state $I_i\in{I_Q}$ and goal $G_i\in{G_Q}$. The exact logical description of these tasks is not known a priori but is conveyed through task-specific demonstrations in the form of example plans. 

A generalized plan for TAMP can be understood as an abstract "program" for solving the problem, which specifies the order and behavior of high-level operations but does not directly specify all of the continuous parameters. Unlike generalized symbolic plans, we cannot execute this program step-by-step, choosing parameters "greedily" as we go, because of additional constraints imposed by the geometry and physics of the environment. For example, greedily sampling and executing placements of objects in a packing task may lead to situations where objects placed earlier in a plan obstruct objects' placements later in the plan. For this reason, the objective of a generalized task and motion planner is to construct a new constrained domain for each task that, when solved using a \tamp{} solver, finds a ground plan most efficiently in terms of wall-clock computation time for any problem instance conforming to the task specification. (Figure ~\ref{fig:concept})

This formulation is notably different from that of generalized planning in a few important ways. First, consistent with \tamp{} robotics applications, all tasks are specified under the same lifted domain containing the same action sets, streams, and properties describing the world. Second, goals and initial states are now defined in a logical language and involve continuous geometric and physical parameters (e.g., a pose is within a region.) Lastly, because direct execution of a generalized \tamp{} plan is impossible, we instead evaluate search speed across several testing problem instances.

\noindent
\textbf{Working example.}
To build intuition for generalized \tamp{}, we describe a simple working example of a generalized \tamp{} problem in terms of the specification described above. First, we establish an example overarching robotics domain containing a ``free-flying'' robotic gripper agent that can pick and place movable objects.

This domain has predicates $\texttt{Holding}(o)$, $\texttt{On}(o_1, p, o_2)$, $\texttt{Movable}(o)$, $\texttt{Tray}(o)$, and $\texttt{HandEmpty}$. The domain also has certified predicates $\texttt{Pose}(o, p, s)$ for possible object placement and $\texttt{CFree}(o_1, o_2, p_1, p_2)$ indicating that two objects at certain poses on the same surface are not in collision. Additionally, the domain will have the following action schemas:

\begin{small} % footnotesize
\begin{lstlisting}
Pick(|$o, p, s$|)
 |{\bf pre}:| HandEmpty, On($o, p, s$), Pose($o, p, s$)
 |{\bf eff}:| $\;$Holding($o$), $\lnot$On($o, p, s$), $\lnot$HandEmpty
Place(|$o, p, s$|)
 |{\bf pre}:| Holding($o$), Pose($o, p, s$),
      $\forall o_2, p_2:\; $CFree($o, p, o_2, p_2$)
 |{\bf eff}:| $\lnot$Holding($o$), On($o, p, s$), HandEmpty
 \end{lstlisting}
\end{small}

Lastly, the domain has two streams that generate sampled entities and certified predicates. The first stream $\texttt{SamplePlacement}(o, s)$  takes in a surface and an object, samples a pose entity $p$, and adds $\texttt{Pose}(o, p, r)$ to the initial state. The second stream  $\texttt{CheckCollision}(o_1, p_1, o_2, p_2)$ tests for collisions and adds the certified predicate $\texttt{CFree}(o_1, p_1, o_2, p_2)$ if none were detected.

We now define the $\textsc{Load}$ task, one of many possible tasks within this domain. The goal of this task ($G_\textsc{Load}$) is for all objects to be on the tray or for the tray to be full, and the initial state ($I_\textsc{Load}$) is all objects on the table.
The $\textsc{Load}$ task consists of problem instances $\{P_N, \dots\}$ where $N$ refers to the number of objects. The initial entitiy set for any problem within this task consists of a robotic gripper agent $r$, a set of $N$ objects $b_1, \dots b_N$ and their corresponding initial poses $\mathit{bp}_1, \dots \mathit{bp}_N$, a tray $t$ with an initial pose $\mathit{tp}$, and a table $\mathit{src}$. The literals in the initial state of a problem instance are $\texttt{HandEmpty}$, $\texttt{Tray}(t),$ $\,\texttt{On}(t, \mathit{tp}, \mathit{src}),$ $\, \texttt{Pose}(t, \mathit{tp}, \mathit{src}) $, $\texttt{On}(b_i, \mathit{bp}_i, \mathit{src})$, $\texttt{Pose}(b_i, \mathit{bp}_i, \mathit{src})$, and $\texttt{Movable}(b_i)$ for all $b_i$. An optimal policy for this simple task is to repeatedly pick up an object and place it on the tray until the tray is full and then pick up the tray. This task is a subtask of the $\textsc{Transport}$ task, which is described and evaluated in the Experiments section.

\section{\textsc{GenTAMP} Training}

% First introduce the overall pipeline and reference Bonet \textbf{once}. Add block diagram.
In this section, we describe our proposed method for learning a deterministic policy from very few small \tamp{} example plans. In the first step, we uses example plans to extract a pool of unique features of states in the domain. Next, we combine these features with a set of constraints to form a SAT theory which, when solved, yields a subset of features from the feature pool that comprise an \textit{approximately valid abstraction}. We then use these selected features and an initial state to construct a fully observable nondeterministic (FOND) planning problem. The solution to the FOND problem is a policy that deterministically returns an abstract action when queried with an evaluation of the selected features. Details of the policy learning pipeline are in Algorithm~\ref{alg1}. This strategy for generalized planning was introduced by ~\citet{learning_abstractions_2}. In the following sections we describe a novel version of this framework that applies to problem domains with both continuous state and action spaces.

\begin{algorithm}
\caption{\textsc{GenTAMP} Training}
\label{alg1}
\KwIn{$P = \{P_1, P_2, \dots, P_{N-1}\}\subseteq{Q}$}
\KwOut{$\pi_\phi$}
\hrulefill\\
\nl $\mathcal{D}$ = \{\}, $\text{pool} \gets \{\}$\\
\nl \For{$P_i \in P$}{ 
\nl $\mathcal{D} \gets \mathcal{D}\;\cup\;\textsc{TAMP}(P_i)$\\
}
\nl \For{$c \in \{1\;\dots\;\text{max\_complexity}\}$}{
\nl $\text{pool}\gets \textrm{\textsc{Prune}}(\text{pool}\; \cup \; \textrm{\textsc{Generate}}(\mathit{pool}))$\\
}
\nl $\mathcal{T}\gets\textsc{EncodeSAT}(\mathcal{D}, \text{pool})$ \\
\nl $F, \bar{A}\gets\textsc{SATSolve}(\mathcal{T})$\\
\nl $\bar{I}, \bar{G}\gets{F(P)}$\\
\nl $\pi_\phi \gets\textsc{FONDPlan}(F, \bar{A}, \bar{I}, \bar{G})$\\
\nl \Return $\pi_\phi$, F
\end{algorithm}

\noindent
\textbf{State and action abstractions.}
To define a policy with inputs conforming to a common representation across problem instances within a task, we establish a transformation of the problem-instance state space. This can be done using a set of Boolean features $\phi_b: \mathcal{S}_i \rightarrow \{0, 1\}$ for any instance state space $\mathcal{S}_i$ and numeric features $\phi_n: \mathcal{S}_i \rightarrow \mathbb{N}_0$ and their corresponding qualitative projections $\phi_{\tilde{n}}: \mathcal{S}_i \rightarrow \{0,\mathbin{>}0\}$  for any instance state space $\mathcal{S}_i$. These features are defined as relations over sets of entities and can therefore be applied to states with varying numbers of entities and properties. For example, the numeric feature $\phi_n(s) \equiv |\{x\in \mathcal{O}_s : \texttt{Blue}(x)\}|$ takes in a state s and returns the number of blue objects in that state. Such a feature can be applied to any state regardless of the number of objects in that state.

$F$ denotes the set of Boolean and qualitative numeric features.  In general, these features are insufficient to reconstruct the state, and so yield an abstract state representation $\phi(s)$, which is simply an evaluation of each $\phi_b$, $\phi_{\tilde{n}}$ for a state from any of the problem instances in the generalized planning problem. This abstract transformation can be applied to the initial and goal states of a problem instance to obtain an abstract initial state $\bar{I}$ and goal state $\bar{G}$. Abstract actions $\bar{A}$ are defined by preconditions and effects that are Boolean features or qualitative evaluations of numeric features. Abstract actions are easily derived from the features $\phi_b$, $\phi_{\tilde{n}}$. Specifically, the effects of a derived abstract action are the qualitative effects over the features, and the preconditions of a derived action are the features that agree qualitatively in the states where that qualitative effect was observed. These abstract actions are no longer deterministic since actions that decrease a numeric feature can have two possible effects on the qualitative abstract state: reducing the feature value to $0$ or leaving it greater than 0.

An abstraction for a task is the tuple $\langle F, \bar{A}, \bar{I}, \bar{G} \rangle$. Given an abstraction, an abstract policy $\pi_{\phi}(s)$ can be found using generate-and-test methods \cite{SZIGaaai11} or, more efficiently, using a non-deterministic planner~\cite{Muise2012ImprovedNP}. This abstract policy can be applied to novel problem instances containing an arbitrary number of objects.

\label{sec:feature_learning}

Our objective is to learn $\phi_b$ and $\phi_n$, and the requisite abstract actions, for a task $Q$ from data representing transitions sampled from problem instances of the task $Q$. The learned feature sets and resulting abstract actions should satisfy three critical properties: soundness, completeness, and goal-distinguishability (see appendix for definitions).

These three properties of the abstraction are proven to be \textit{valid}, in the sense that they are sufficient to guarantee that a policy in the space of abstract actions and states will always be refinable to a concrete plan in any problem instance of the domain~\cite{learning_abstractions_1}.

\noindent
\textbf{Collecting training data.}
\label{sec:data_collection}
Generating valid abstractions requires an exhaustive search through the state space to verify soundness, completeness, and goal distinguishability on every transition. However, due to the continuous, high-dimensional, and nonlinear nature of \tamp{} problems, it is impossible to enumerate every reachable state and transition in the environment. For this reason, we focus on constructing
\textit{approximately valid} abstractions from a subset of possible transitions in the environment.

Our approach uses a \tamp{} solver to find example plans for each training problem instance and uses the transitions from those plans to create approximately valid abstractions. The constituent states of the plan consist of all symbolic predicates and entities along with any sampled entities and certified predicates that are in the preimage of the plan.

\noindent
\textbf{Generating a feature pool.}
To select a set of features that constitute an approximately valid abstraction, we generate a pool of complexity-bounded candidate features and select a subset of those features that jointly yield a valid abstraction. To accommodate high-arity certified predicates with sampled entities, we design a generative grammar that enables the use of features containing implications and mixed existential and universal quantifiers.

A \tamp{} domain comes with a set of primitives predicates $p(\bar{o})$. We additionally establish a set of named variables $\{x_1, \dots , x_M\}$ where $M$ is the maximum predicate arity. These predicates and named variables are used as terminals in forming a set of concepts $C(\bar{z})$ through primitive negation, conjunction, and a single implication. Concepts are combined to form quantified concepts $C_q(x, \bar{z}_1, \bar{z}_2)$ where $x$ is a free variable, $\bar{z}_1$ is a tuple of bound variables, and $\bar{z}_2$ is a tuple of variables that have yet to be bound. Formula containing only a single free variable are formed by quantifying all but one of the free variables using chained universal and existential quantifiers. Lastly, features are formed from the quantified concepts that contain only a single free variable. The resulting features are functions that take in a state and return a natural number that specifies the number of entities that can be plugged into the remaining free variable such that the feature's concepts holds. The specifics of this generative grammar are in Table~\ref{table:pcfg}. 

Each feature in this grammar is accompanied with a \textit{feature complexity} score that is defined as the maximum of the number of generative grammar rules applied to create the feature and the number of arguments of the feature. This grammar is implemented in a bottom-up fashion by combining primitives, concepts, and quantified concepts to create a full feature and discarding features when they pass a fixed complexity limit (5 in our experiments).

\renewcommand{\arraystretch}{1.3}
\begin{table}[h]
\begin{center}
\begin{tabular}{| l |}
\hline
\textbf{Arguments}  \\
$z_m \rightarrow x_1\;|\; x_2\;|\; \dots\;|\; x_M $\\ 
\hline
\textbf{Concepts}  \\
$C_r(\overline{z}) \to  p(\overline{z})$\\
$C_r(\overline{z}) \to  \lnot\;p(\overline{z})$\\
$C_r(\bar{z}_1\circ\bar{z}_2) \rightarrow C_{r_1}(\bar{z}_1) \land C_{r_2}(\bar{z}_2)$\\
$C \rightarrow C_{r}(\bar{z})$ \\
$C \rightarrow \;C_{r_1}(\bar{z}_1) \Rightarrow C_{r_2}(\bar{z}_2)$\\
\hline
\textbf{Quantified Concepts}  \\
$C_q(x, \emptyset, \bar{z}) \rightarrow C(\{x\}\circ\bar{z}) $\\
$C_q(x, \bar{q}\circ{\{z_1\}}, \bar{z}_{2:N} ) \rightarrow \exists{z_1}. C_q(x, \bar{q}, \bar{z}_{1:N}) $\\
$C_q(x, \bar{q}\circ{\{z_1\}}, \bar{z}_{2:N} ) \rightarrow \forall{z_1}. C_q(x, \bar{q}, \bar{z}_{1:N})$ \\
\hline
\textbf{Features}  \\
$\phi_n \rightarrow f: s\mapsto\vert \{x \in \mathcal{O}_s : C_q(x, \bar{z}, \emptyset)\}\vert$\\
\hline
\end{tabular}
\caption{Generative grammar for features that take in a state and generate per-entity first-order logical expressions.}
\label{table:pcfg}
\end{center}
\end{table}

% \lpknote{We should probably talk this grammar through----I have some questions about it.}

% \arcnote{Do I need to describe how this grammar is actually implemented? -- It's not the same as how it is described here for efficiency reasons}

%\begin{lstlisting}
%C($x$) $\gets \lnot $C($x$)
%C($x$) $\gets$ C($x$) $\land $ C'($x$)
%C($x$) $\gets \exists \bar{y} $(R($x; \bar{y}$) $\land$ C($x$))
%C($x$) $\gets \forall \bar{y} $(R($x; \bar{y}$) $\land$ C($x$))
%C($x$) $\gets \exists z\oplus\bar{y} $(R($z; \bar{y}$) $\land$ C($x$))
%C($x$) $\gets \forall z\oplus\bar{y} $(R($z; \bar{y}$) $\land$ C($x$))
%R($x; \bar{y}$) $\gets\lnot$ R($x;\bar{y}$)
%R($x; \bar{y}$) $\gets$ R(head$(\bar{y})$; tail($\bar{y}$)$\;\oplus\;$x)
%F $\gets$ F $\cup$ $f: s\mapsto\vert \{x \in \mathcal{O} \;\;\vert$ %C($x;s$)$\}\vert$
%\end{lstlisting}

% \tsnote{what is the $C$ in $x_C$?}
% \tsnote{I'm guessing that overline means zero or more terms, worth defining}
% \tsnote{what are the $P$'s here? they're different from the $P_i \in Q$ training problems right? and what are their subscripts?}
% \tsnote{same question about $Q$}

This grammar specification has two important properties necessary for \tamp{} domains. First, we use arbitrary-arity features instead of binary features to accommodate the high-arity predicates present in most \tamp{} domains. To handle these higher arity predicates, our grammar includes alternating quantifiers evaluated in Prolog~\cite{prolog}. Second, we incorporate certified predicates and sampled entities into the grammar through the terminals ($p$) so that the resulting features are mixtures of geometric and symbolic properties. These changes greatly increase the total number of features and thus require more aggressive pruning at each generation step. We prune concepts and quantified concepts in each step by $\textit{uniqueness}$ and $\textit{equivalence}$. Uniqueness pruning looks at each collected transition and checks to ensure no lower-complexity feature has the same value on each transition. Equivalence pruning uses standard logical inference rules to test equivalence between logical formulas.

Listed below are some example quantified concepts from our working $\textsc{Load}$ task, including quantified concepts composed from both symbolic and geometric primitives from the \tamp{} specification.

\begin{itemize}
  \item $f_1(o) \equiv \texttt{Holding}(o)\;\land\;\lnot\texttt{Tray}(o)$
  \item $f_2(o) \equiv \texttt{Holding}(o)\;\land\;\texttt{Tray}(o)$
  \item $f_3(o) \equiv \exists{p, r}:\texttt{On}(o, p, r) \land \lnot\texttt{Tray}(r)$
  \item $f_4(o) \equiv \exists{p, r}\forall{o_2, p_2}:\texttt{Tray}(r) \land \lnot \texttt{On}(o, p, r) \land \texttt{On}(o_2, p_2, r) \Rightarrow \lnot \texttt{CFree}(o, p, o_2, p_2, r)$
\end{itemize}

When these quantified concepts are converted to features, we get numeric evaluations of each concept. Using the final concept as an example, the feature corresponding to this concept returns the number of objects that can be placed on a tray without collision with existing objects on the tray. This feature combines sampled continuous properties of the state, such as collision constraints, with symbolic properties such as $\texttt{On}$ and $\texttt{Tray}$. It is easy to imagine how this feature might be useful in a generalized task and motion plan for the $\textsc{Load}$ task.

\noindent
\textbf{Abstraction Learning.}
\label{sec:sat}
Given a feature pool generated by the above grammar and the set of collected transitions, we now try to find a subset of that feature pool that satisfies the soundness, completeness, and goal-distinguishability criteria for the example plans in the training dataset. We encode the desired constraints (outlined in the appendix) as a SAT theory, and then apply a SAT solver to generate an approximately valid abstraction. Our implementation uses the OpenWBO ~\cite{openwbo} weighted Max-SAT solver and sets weights on the features that correspond the the complexity of that feature in the grammar. After applying the SAT solver to our working $\textsc{Load}$ task to enforce our criteria over the set of selected features, we find that that the lowest-weight set of features satisfying the correctness criteria is exactly those specified in the previous section; the extracted actions are in Figure~\ref{fig:example}. This is typically the computationally slowest part of the policy learning pipeline because constructing the SAT theory requires evaluation of every numeric feature on all example transitions (see Table 1 in the appendix).

Because of the non-closed world assumption of sampled entities and certified predicates, we cannot guarantee the correctness of any feature evaluation on the state that quantifies over continuous state variables. For example, we cannot assert the correctness of the Boolean feature ``All object grasps are not reachable'' because there are an infinite number of possible grasps. However, assuming the sampled entities are drawn uniformly from their respective continuous spaces, the features will be evaluated correctly in the limit of infinite samples. Because this uniform assumption is violated when extracting sampled entities in the preimage of \tamp{} plans, we add additional stream samples in each plan prior to abstraction learning. Details of this procedure are in the appendix. 

\noindent
\textbf{Qualitative Numeric Planning.}
Given an abstraction, we apply the feature transformation to the first and last states of the training instances to get abstract representations of our initial state and goal. This abstract fully observable non-deterministic planning (FOND) problem is then passed to a standard non-deterministic planner to find a generalized plan. In this work, we use state-of-the-art PRP~\cite{Muise2012ImprovedNP} as our FOND planner. Figure~\ref{fig:example} shows the resulting generalized plan for the working \textsc{Load} task given our learned features.

\section{Using a Generalized Task and Motion Plan}
\label{sec:algorithm}

We now discuss how the learned generalized plan can be concretized (or executed) in a novel \tamp{} problem instance within the same task. Typical approaches to generalized planning execute learned policies by selecting randomly from a set of ground actions that 1.) are feasible in the state and 2.) induce the same effects on the abstract state as the abstract action from the policy. However, direct execution of a generalized task and motion plan is impossible due to additional geometric and physical constraints induced by the environment.

% An important note is that, due to the formulation of the TAMP problem, geometry and other physical properties of the environment can only \em{add} constraints in the form of additional arguments in the action preconditions. That is, geometric properties of the environment never create affordances which are otherwise not present in the purely symbolic instantiation of a task. For this reason 
% \tsnote{this brings up the question of what theoretical guarantees do you get from the learned abstract policy. I'm guessing that with some assumptions, greedily executing abstract policies in discrete gentamp is guaranteed to succeed. so that isn't true here, but is there a guarantee that the abstract policy leads to a skeleton that can be refined, or anything like that?}

One common structure for \tamp{} planners is to have an outer loop that searches over plan ``skeletons'', which consist of operator instances with the discrete parameters bound, but with the continuous parameters still free, and a set of constraints that must be satisfied in order to make the plan valid.  Then, an inner loop searches for a satisfying assignment of the continuous parameters. To embed our learned policy into the standard \tamp{} framework, we construct a new set of planning operators that encodes the action-ordering constraints and abstract action preconditions and effects of the generalized plan, and feed these new operators into \pddlstream{}. The resulting constrained domain can perform a more efficient concurrent search through the joint space of skeletons and continuous-parameter bindings, thus reducing the overall search time of the \tamp{} solver. The details of this algorithm are in Algorithm 1 of the appendix. 

Using our $\textsc{Load}$ task as an example, while a traditional \tamp{} solver would explore actions that place objects on the table, this would be strictly infeasible in the constrained task-specific problem specification because the abstract $\textsc{Load}$ policy specifies that if an object is being held, the next action must increase the number of objects on the tray (Figure~\ref{fig:example}). 

\section{Experiments}
\label{sec:experiments}

\begin{figure*}[h]
\centering
        \includegraphics[width=\linewidth]{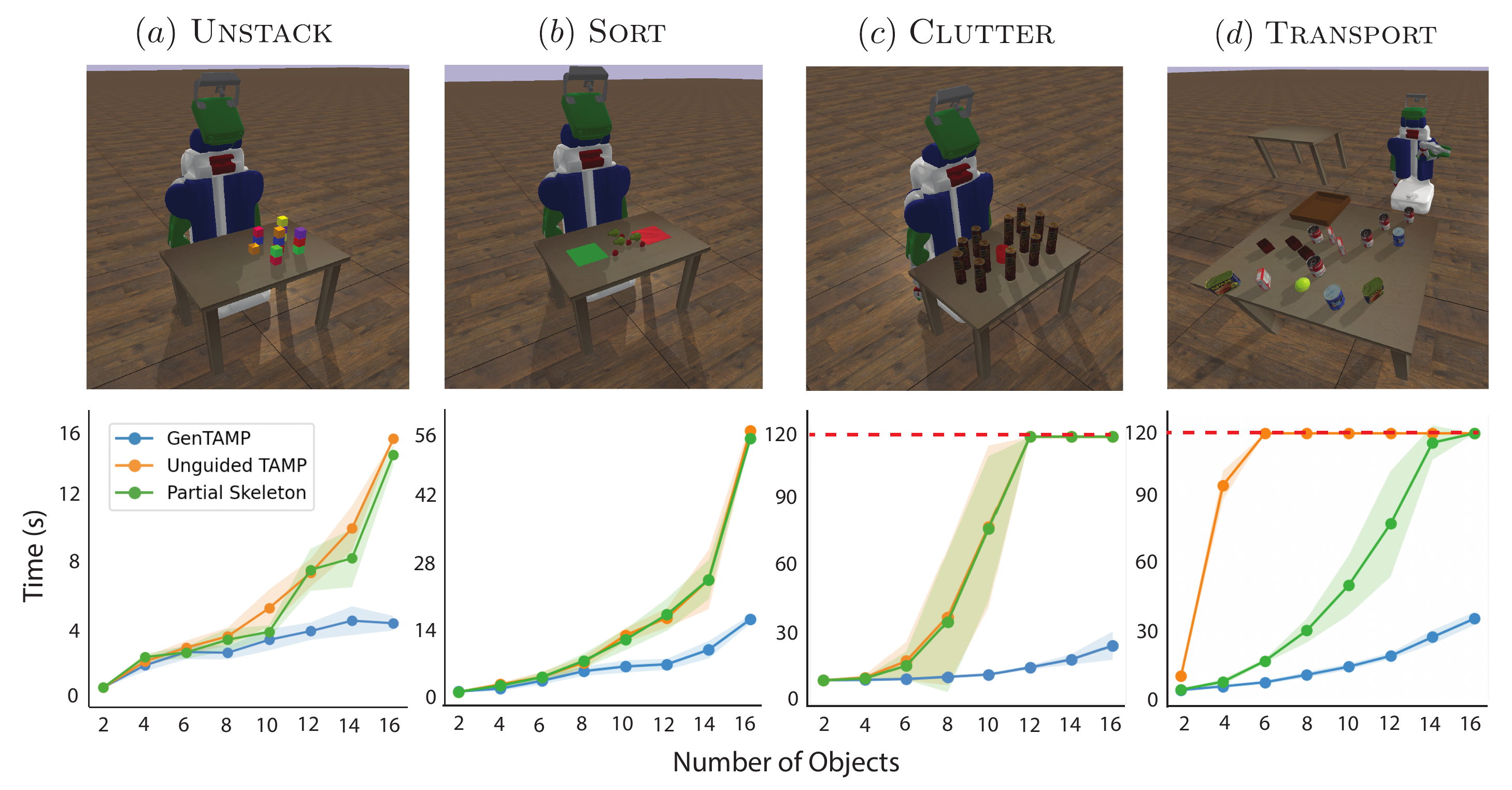}
    \caption{The top panels show sample initial states for our continuous \tamp{} problems. The bottom four panels show planning speed comparisons between our approach, standard \tamp{}, and partial skeleton. Each trial contains six seeds used to randomize object type, placement, size, and color. The plot error bars show a 95\% confidence interval for planning time. The dotted red line indicates a 120 second timeout.
    % \arcnote{Make images brighter. Zoom in on the task. Add GNN Continuous BC baseline.}
    }
    \label{fig:tasks}
\end{figure*}

We demonstrate \textsc{GenTAMP} on four continuous robotic \tamp{} tasks described below and visualized in Figure~\ref{fig:tasks}. We chose tasks that are physical instantiations of widely used tasks in generalized planning literature ~\cite{Srivastava2011ANR, learning_abstractions_2, Illanes_McIlraith_2019} or classically difficult \textsc{TAMP} domains with large numbers of objects. Here we describe the goal of each of the tasks and the qualitative behavior of the policies discovered by \textsc{GenTAMP}.

We compare \textbf{\textsc{GenTAMP}} with two baseline approaches. \textbf{Unguided \textsc{TAMP}} is simply an execution of a state-of-the-art task and motion planner (the same one used for data collection). \textbf{Partial skeleton} uses the same task and motion planner but provided with an input skeleton that contains a known feasible action schema ordering for each task. The actions in the skeleton do not have populated object parameters, reducing the problem to constraint satisfaction. We consider four tasks within a domain containing a PR2 robot with full kinematic constraints. 

\noindent
\textbf{Unstack:} The objective of this task is to remove all of the stacked blocks and place them on the surface of the table. The policy discovered by \textsc{GenTAMP} is to find a block with no other block stacked on top of it, pick it up, and place it on the table until no stacked blocks remain.

\noindent
\textbf{Sort:}
The objective of this task is to place all of the objects on the table into their similarly colored regions. The policy discovered by \textsc{GenTAMP} is to place all the green objects in the green region, followed by placing all red objects on the red region. This policy is discovered despite none of the example plans sorting in this color-specific order.

\noindent
\textbf{Clutter:}
The goal is to pick up the target object, but this is made difficult by obstruction from many distractor objects. The policy discovered by \textsc{GenTAMP} is to identify the objects it collides with when reaching for the target object and move those objects out of the way by placing them elsewhere on the table until it can reach the target object. This task is known to be difficult for \tamp{} solvers and has previously required significant engineering, including collision error feedback, to be feasible in large problems \cite{srivastava_clutter}. \textsc{GenTAMP} avoids such engineering by learning features from small examples that guide search toward moving objects that block trajectories.

\noindent
\textbf{Transport:} The objective of this task is to carry all of the objects on the source platform to the destination platform. The agent has access to a tray that it can use to carry over multiple objects to the destination region. The strategy discovered by \textsc{GenTAMP} is to place objects onto the tray until there is no more physical space on the tray, carry the tray to the destination location, and unload the tray until no objects remain on the source location. Because objects are different sizes, the generalized \tamp{} plan reasons about geometry to determine if more objects can be placed on the tray.

As seen in Figure~\ref{fig:tasks}, our experiments show that execution speed from the $\textsc{GenTAMP}$ policy greatly outperforms a state-of-the-art \tamp{} algorithm. The difference in planning speed increases exponentially with object count. Visualization of the topology for each of the tasks and the learned abstractions along with further experimental details can be found in the appendix.

\section{Related work}
\label{sec:related}

Generalized planning for purely discrete domains has been the subject of much prior work~\cite{survey}. 
One of the earliest approaches to generalized planning was a generate-and-test formulation where plans are found for single problem instances and tested on others~\cite{kplanner,fsa_planner}. An alternative approach is to iteratively merge example plans into a single generalized plan~\cite{Srivastava2011ANR, Srivastava2011DirectedSF}. Generalized planning has also been accomplished using heuristic-based solvers that search through the entire space of generalized plans ~\cite{Lotinac2016AutomaticGO, hlsf_generalized_planning, FSM_GP}. Some more recent work finds generalized plans in symbolic domains by searching through a space of learned features and abstractions to find successful policies~\citep{learning_abstractions_1, learning_abstractions_2, learning_abstractions_3}. Our \gentamp{} approach most closely follows this strategy~\cite{learning_abstractions_1} with a focus on applications to continuous geometric domains through the use of a continuous \tamp{} solver.

% Generalization in robotic reinforcement learning
There have been various attempts at using offline reinforcement learning and imitation learning to discover useful policies for robotic agents from example traces when no domain description is available. However, reinforcement learning typically struggles with sample complexity and generalizability to domains not seen during training, requiring thousands of example interactions to learn a very specific policy. While some work has been done on finding generalizable policies, these generalization capabilities are limited to visual changes or changes to the dynamics of the simulated environments~\cite{background_rl, background_change, Peng_2018}. To our knowledge, there is no existing imitation learning or offline reinforcement learning method that generates policies for robotics domains that can learn from a few examples and generalize to arbitrary numbers of objects and geometric configurations.

% Learning to accelerate task and motion planning
The problem of accelerating \tamp{} or planning in general by learning from previous planning solutions is an active area of research.
One class of methods accelerates \tamp{} by reducing the size of the search space in various ways. Such methods include identifying goal-irrelevant objects during search~\cite{ploi}, focusing the search on certain regions of the state space by learning constraints~\cite{camps}, performing learned partial grounding~\cite{partial_action_grounding} of operators.
Other methods attempt to accelerate \tamp{} by biasing the sampling procedures embedded within the continuous \tamp{} algorithm by learning scoring functions over the sample space~\cite{score_space}. 
We aim to speed up \tamp{} in tasks where all objects, operators, and state-space regions are potentially relevant without biasing the internal sampling strategies. 

\section{Conclusion} 

% There are many ways of generating subsets of transitions from large environments that lead to various approximately valid abstractions. Random action sampling from the initial state will result in a subset of states in the environment, but if the goal is unlikely to be reached through random actions, there will be no data to verify whether the goal-distinguishability constraint was satisfied. A more principled approach, which we employ in this paper, is to collect example transitions $(s, s')$ from satisficing plans to a set of small training problems $P_i \in Q$. One limitation of this approach is the lack of negative examples to constrain the features which distinguish goal states. A possible extension of this approach involves sampling ``around'' a satisficing plan to produce negative goal examples around the goal state.

In this paper, we introduced a new generalized task and motion planning problem definition inspired by a desire to find generalizable and scalable policies for continuous robotics tasks. We designed an algorithm for solving this generalized \tamp{} problem by learning task-specific abstract policies that apply to problems with arbitrary numbers of objects and initial states from only a few example plans. Lastly, we showed that these simple policies increase the search speed of \tamp{} solvers in large domains, leading to more scalable and generalizable long-horizon robotic planning.
\label{sec:conclusion}

\section*{Acknowledgments}
We gratefully acknowledge support from NSF grant 1723381; from AFOSR
grant FA9550-17-1-0165; from ONR grant N00014-18-1-2847; from the Honda Research
Institute; and from MIT-IBM Watson Lab. Aidan Curtis and Tom Silver are supported by NSF GRFP fellowships. 
Any opinions, findings, and conclusions or recommendations expressed in this material are those of the authors and do not necessarily reflect the views of our sponsors.

\bibliography{arxiv}
\pagebreak

\section{Appendix}

\begin{algorithm}[h]
\caption{\textsc{GenTAMP} Policy Execution}
\label{alg:embedding_and_execution}

\KwIn{$\langle \mathcal{S}_{N}, \mathcal{I}_{N}, A_{N}, G_{N} \rangle = P_{N}, \;\pi_{\phi}, F$}
\KwOut{A valid \tamp{} plan: $(a_1, s_1), (a_2, s_2), \dots, (a_n, s_n)$}
\hrulefill\\
\nl plan $, NA_{N} \gets []$, $\mathcal{I}_a \gets F(\mathcal{I})$, Queue $\gets [\mathcal{I}_a]$\\
\nl $\forall{\phi \in F}: Axioms[\phi] \gets \text{GenerateAxiom}(\phi)$ \\
\nl \While {Queue} {
    \nl $S_a \gets Queue.pop()$\\
    \nl $A_a \gets{\pi_{\phi}(S_a)}$\\
    \nl $S'_a \gets{apply(S_a, A_a)}$ \\
    \nl \For{$A_i \in A_N$}{
            \nl NewAction = $\text{copy}(A_i)$\\
            \nl $\forall{\phi \in A_a.pre}:$ NewAction.add\_pre(Axioms[$\phi$])\\
            \nl $\forall{\phi \in A_a.eff}:$ NewAction.add\_eff(Axioms[$\phi$])\\
            \nl NewAction.add\_pre(Order($S_a$)) \\
            \nl NewAction.add\_eff($\lnot$ Order($S_a$)) \\
            \nl NewAction.add\_eff(Order($S'_a$)) \\
            \nl Add NewAction to $NA_{N}$
    }
    \nl Queue.add($S'_a$)\\
}
\nl I.add\_part(Order$(I_a)$)\\
\nl plan = TAMP$(\langle \mathcal{S}_{N}, \mathcal{I}_{N}, NA_{N}, G_{N}\rangle)$\\
\nl \Return $\text{plan}$\\
\vspace{.3cm}

\end{algorithm}

\subsection{Experimental Details}
\textsc{GenTAMP}, Unguided \tamp{}, and Oracle Skeleton are all evaluated with \pddlstream{} \cite{pddlstream} using the same \pddlstream{} search control strategy and greedy best-first search solver using the h$_{\text{Add}}$ heuristic. Axioms are evaluated using SWI-Prolog \cite{swipl}. All abstraction learning is done with a maximum feature complexity of 5. All experiments were run in a Docker container in Google Cloud Kubernetes Engine (GKE) on an e2-standard-16 machine, with each container having access to 1 CPU and 4 GB of memory. Our results in the experiments section only show \textsc{GenTAMP} execution time after the policy was learned. A runtime breakdown of various components of policy learning can be found in Table~1 in the appendix.

\subsection{Valid Abstraction Definitions}

\begin{definition}
An abstract action is \textbf{sound} over the features $\phi$ if for each possible transition $(s, s')$ in each problem instance $P_i\in Q$ where the abstract action preconditions hold in $s$, there exists a concrete action with the same qualitative effects on the abstract state, that is, that $\exists a. \phi(f(s, a)) = \phi(s')$.
\end{definition}

\begin{definition}
A set of abstract actions is \textbf{complete} in $Q$ if for any concrete state $s$ in any $P_i\in{Q}$ where there is an applicable concrete action in $s$, there is also an applicable abstract action in $\phi(s)$.
\end{definition}
% \lpknote{Is there an important correspondence?  That is, does the concrete action have to correspond in some sense to the abstract one?}

\begin{definition}
A feature set is \textbf{goal-distinguishable} in $Q$ if, given any pair of states $(s, t)$ where $s$ is a non-goal state and $t$ is a goal state, there exists at least one feature $\phi_i$ (Boolean or qualitative numeric) such that $\phi_i(s)\neq\phi_i(t)$.
\end{definition}

\begin{definition}
An abstraction $\langle{F, \bar{A}}\rangle$ is \textbf{valid} if its abstract actions $\bar{A}$ are sound and complete and its feature set $F$ is goal-distinguishable.
\end{definition}

\subsection{SAT Theory Description}
The SAT Theory used for this paper are the same as those used in ~\citet{learning_abstractions_2}. In the following theory $\mathcal{D}$ represents the database of transitions and $\mathcal{G}$ represents the subset of transitions that are goal-satisfying. The output of the SAT theory is a binary classification ($select$) associated with each feature that denotes whether or not that feature is selected for the abstraction.\\

\noindent
\textbf{Variables}
\begin{itemize}
\item $select(f)$
\item $D_1(s, t)$
\item $D_2(s, s', t, t')$
\end{itemize}
\noindent
\textbf{Constraints}
\begin{itemize}
\item $D_1(s, t) \iff \bigvee\limits_{\{f\in{F}| f(s) \neq f(t)\}}select(f)$
\item $D_2(s, s', t, t') \iff \bigvee\limits_{\{f\in{F}| \Delta_f(s, s') \neq \Delta_f(t, t')\}}select(f)$

\item $\forall \lnot{D_1}(s, t) \implies \lnot D_2(s, s', t, t')$
\item $\forall t\in \mathcal{G}, s\notin \mathcal{G} \;\; {D_1}(s, t)$
\end{itemize}
\begin{figure}
\centering
    \includegraphics[width=\linewidth]{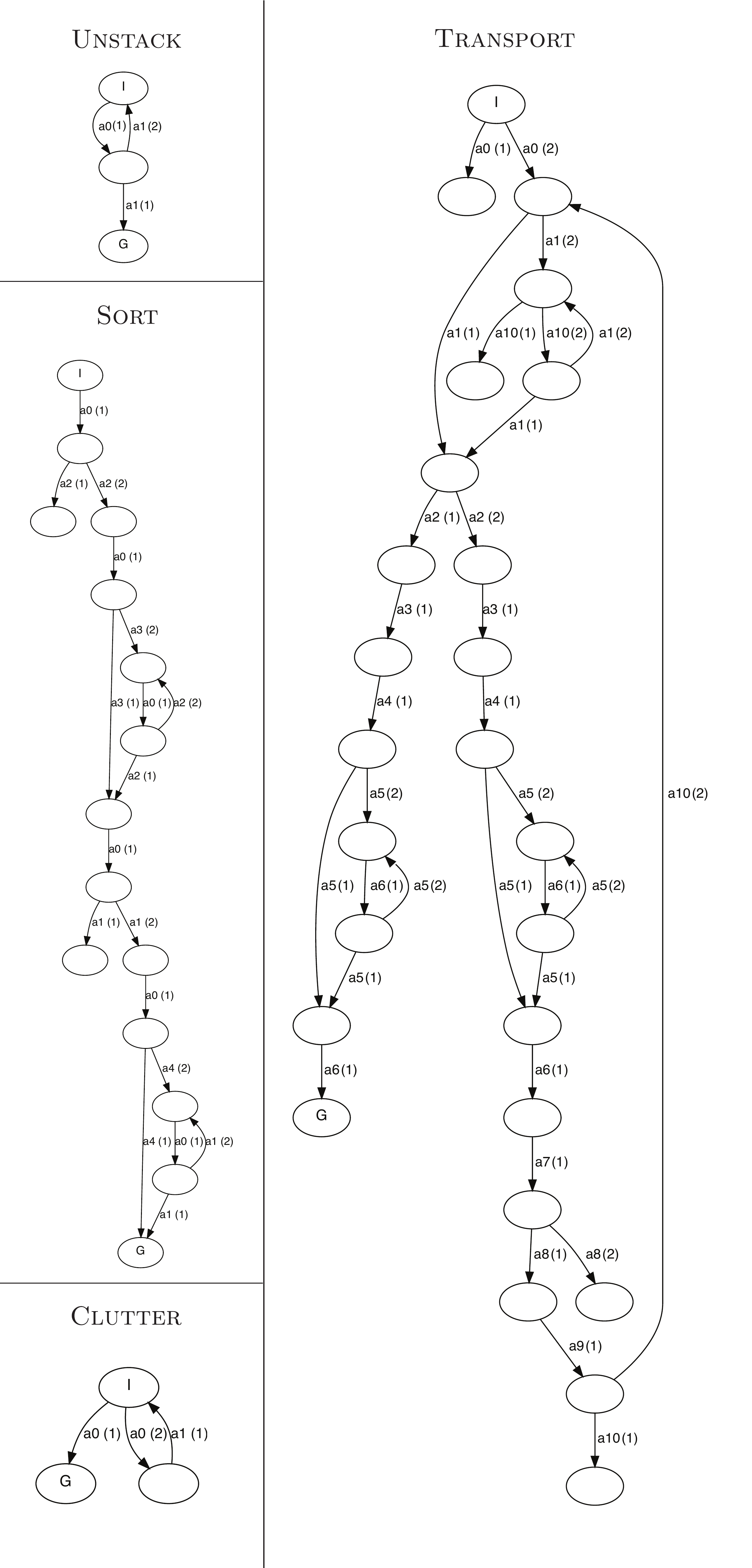}
    \caption{The abstract policies for each task extracted from the learned abstractions via FOND planning. The action names on the edges correspond to the actions outlined in the abstraction descriptions in the appendix.}
\end{figure}

\subsection{Training/Testing Problems}
Below we list the specifics of the testing and training problems used in our experiments.\\

\noindent
\textbf{Unstack:}
\noindent
Blocks were placed sequentially on either the table or on another already placed block. The choice of surface to place the next block was chosen uniformly from the available clear surfaces and the table. When placed on the table, the pose was selected uniformly over the space of stable placements and checked for collisions with other objects. The training problem contained 8 blocks and the testing problems contained between 2-16 blocks.

\noindent
\textbf{Sort:}
Objects of both colors were place with random poses such that the y value of that pose fell between the two colored regions. Sampled were checked for collisions with other objects. The training set were of the form Sort-X-Y where X represented the number of red objects and Y represented the number of green objects. The training set contained Sort-1-1, Sort-2-1, Sort-2-2, Sort-2-3, and Sort-4-3. The testing set contained Sort-A-A with A$\in\{1,\dots,8\}$.

\noindent
\textbf{Clutter:}
A single mug was placed in the center of the table and distractor objects were placed with uniform pose on the table surface while avoiding collisions. Problem instances were of the form Clutter-X where X represented the number of distractor objects. We also verified that, given a joint placement of the distractor objects, $\text{X}//4$ of the distractor objects collided with a sampled trajectory to the target object. This constraint ensures similar complexity for similarly named test problems. The training problems used were Clutter-2, Clutter-5, Clutter-8
and the testing problems were Clutter-2 to Clutter-16 in increments of 2.

\noindent
\textbf{Transport:}
All YCB objects and the tray were placed with random poses on the source table such that none of the objects were in collision. Problem instances were of the form Transport-X-Y where X represented the number of objects that could fit on the tray and Y represented the number of YCB objects initially on the table. The size of the tray was set such that only X objects could be placed on the tray at once. The training problems used were Transport-1-4, Transport-2-4, Transport-3-6. The testing problems used were Transport-2-2, Transport-2-4, Transport-2-6, Transport-2-8, Transport-2-10, Transport-2-12, Transport-2-14, Transport-2-16.

\begin{table*}[]
\centering
\begin{tabular}{|l|l|l|l|l|}
\hline
\multicolumn{1}{|r|}{} & Unstack			        & Sort				    & Clutter			& Transport        \\ \hline
Generate Feature Pool  & $2.77\pm{0.11}$	        & $8.67\pm{0.26}$	    & $1.50\pm{0.51}$	& $0.46\pm{0.03}$  \\ \hline
Create Max-SAT Theory  & $109.49\pm{4.15}$	        & $1025.35\pm{25.94}$   & $202.41\pm{49.62}$& $60.79\pm{1.76}$ \\ \hline
Solve SAT              & $6.24\pm{0.02}$	        & $7.39\pm{0.10}$	    & $6.13\pm{0.03}$	& $7.50\pm{0.12}$  \\ \hline
Solve FOND             & $1.17\pm{0.04}$	        & $1.32\pm{0.34}$	    & $1.13\pm{0.01}$	& $1.13\pm{0.02}$  \\ \hline
Policy Compilation     & $0.01\pm{0.00}$	        & $0.04\pm{0.00}$	    & $0.16\pm{0.01}$	& $0.02\pm{0.00}$  \\ \hline
Total         & $119.7\pm{4.27}$  & \textbf{$1042.75\pm{26.08}$} & \textbf{$211.34\pm{49.99}$} & \textbf{$69.91\pm{1.74}$}     \\ \hline
\end{tabular}
\caption{Breakdown of policy learning speed from various components of the algorithm over six random seeds. All times are in seconds.}
\label{table:policy_learning_speed}

\end{table*}

\subsection{Learned Abstractions}
\noindent
\textbf{Unstack Features}
\begin{itemize}
\item $\textbf{F1}: \texttt{Holding}(y0)$
\item $\textbf{F2}: \forall.y2\;\exists.y1\;  \texttt{On}(y0, y1) \land \texttt{Grasp}(y1, y2)$
\end{itemize}

\noindent
\textbf{Unstack Abstract Actions}

\begin{itemize}
\item $\textbf{Action 0}$\\
      $\textit{pre: }\text{F1}=\bot, \text{F2}>0$\\
      $\textit{eff: }\text{F1}=\top, \text{F2}\downarrow$
\item $\textbf{Action 1}$\\
      $\textit{pre: }\text{F1}=\top$\\
      $\textit{eff: }\text{F1}=\bot$
\end{itemize}

\noindent
\textbf{Sort Features}
\begin{itemize}
\item $\textbf{F1}: \texttt{Holding}(y0)$
\item $\textbf{F2}: \exists.y2\;\texttt{Red}(y0) \land \texttt{On}(y0, y2) \land \texttt{Red}(y2) \land \texttt{Region}(y2)$
\item $\textbf{F3}: \exists.y2\;\texttt{Blue}(y0) \land \texttt{On}(y0, y2) \land \texttt{Blue}(y2) \land \texttt{Region}(y2)$
\item $\textbf{F4}: \forall.y3\;\exists.y0\;\exists.y2\;\texttt{Red}(y0) \land \texttt{On}(y0, y2) \land \texttt{Red}(y2) \land \texttt{Region}(y2) \land \texttt{Grasp}(y0,y1) \land \lnot \texttt{AtPose}(y0, y3)$
\item $\textbf{F5}: \forall.y3\;\exists.y0\;\exists.y2\;\texttt{On}(y0,y1) \land \texttt{Blue}(y0) \land \texttt{On}(y0, y2) \land \texttt{Blue}(y2) \land \texttt{Region}(y2) \land \lnot \texttt{AtPose}(y0 , y3)$
\item $\textbf{F6}: \exists.y1\;\exists.y2\;\texttt{Red}(y0) \land \texttt{Grasp}(y0, y1) \land \lnot (\texttt{Red}(y0) \land \texttt{On}(y0, y2) \land \texttt{Red}(y2) \land \texttt{region}(y2))$
\item $\textbf{F7}: \exists.y1\;\exists.y2\;\texttt{Blue}(y0) \land \texttt{Grasp}(y0, ?y1) \land \lnot (\texttt{Blue}(y0) \land \texttt{On}(y0, y2) \land \texttt{Blue}(y2) \land \texttt{Region}(y2))$
\end{itemize}

\noindent
\textbf{Sort Abstract Actions}
\begin{itemize}
\item $\textbf{Action 0}$\\
      $\textit{pre: }\text{F1}=\bot$\\
      $\textit{eff: }\text{F1}=\top$
\item $\textbf{Action 1}$\\
      $\textit{pre: }\text{F1}=\top, \text{F7} > 0$\\
      $\textit{eff: }\text{F1}=\bot, \text{F3}\uparrow, \text{F7}\downarrow$
\item $\textbf{Action 2}$\\
      $\textit{pre: }\text{F1}=\top, \text{F6} > 0$\\
      $\textit{eff: }\text{F1}=\bot, \text{F2}\uparrow, \text{F6}\downarrow$
\item $\textbf{Action 3}$\\
      $\textit{pre: }\text{F1}=\top, \text{F2}>0, \text{F4}=0, \text{F6} > 0$\\
      $\textit{eff: }\text{F1}=\bot, \text{F2}\uparrow, \text{F4}\uparrow, F6\downarrow$
\item $\textbf{Action 4}$\\
      $\textit{pre: }\text{F1}=\top, \text{F2} > 0, \text{F3} > 0, \text{F5}=\bot, \text{F7} > 0$\\
      $\textit{eff: }\text{F1}=\bot, \text{F3}\uparrow, \text{F5}=\top, \text{F7}\downarrow$
      
\end{itemize}

\noindent
\textbf{Clutter Features}

\begin{itemize}
\item $\textbf{F1}: \texttt{Holding}(y0)$
\item $\textbf{F2}: \forall.y1\;\forall.y2\;\lnot (\texttt{AtPose}(y2, y1) \land \texttt{TrajTo}(y2,y1,y0) \land \texttt{Red}(y2)) \Rightarrow \texttt{HandEmpty}$
\end{itemize}

\noindent
\textbf{Clutter Abstract Actions}
\begin{itemize}
\item $\textbf{Action 0}$\\
      $\textit{pre: }\text{F1}=\bot, \text{F2}>0$\\
      $\textit{eff: }\text{F1}=\top, \text{F2}\downarrow$
\item $\textbf{Action 1}$\\
      $\textit{pre: }\text{F1}=\top, \text{F2}>0$\\
      $\textit{eff: }\text{F1}=\bot, \text{F2}\uparrow$
\end{itemize}

\noindent
\textbf{Transport Features}
\begin{itemize}
\item $\textbf{F1}: \exists.y1\;\forall.y2\;\forall.y3\;\texttt{Pose}(y0, y1)\Rightarrow\lnot\texttt{CFree}(y0,y1,y2,y3) \land \texttt{AtPose}(y2, y3) \land \texttt{Pose}(y2, y3)$
\item $\textbf{F2}: \texttt{Holding}(y0)$
\item $\textbf{F3}: \exists.y1\;\texttt{On}(y0, y1) \land \texttt{Stove}(y1)$
\item $\textbf{F4}: \exists.y1\;\texttt{On}(y0,y1) \land \texttt{Tray}(y1)$
\item $\textbf{F5}: \exists.y1\;\texttt{Holding}(y0) \land \texttt{Pose}(y0, y1) \land \texttt{Tray}(y1)$
\item $\textbf{F6}: \texttt{AtBConf}(y0) \land \texttt{AtStove}(y0)$
\end{itemize}

\noindent
\textbf{Transport Abstract Actions}
\begin{itemize}
\item $\textbf{Action 0}$\\
      $\textit{pre: }\text{F1} > 0, \text{F2} = \bot, \text{F3} > 0, \text{F5} = \bot, \text{F6} = \top$\\
      $\textit{eff: }\text{F2} = \top, \text{F3} \downarrow, \text{F5} = \top$
\item $\textbf{Action 1}$\\
      $\textit{pre: }\text{F1} > 0, \text{F2} = \top, \text{F3} > 0, \text{F5} = \top, \text{F6} = \top$\\
      $\textit{eff: }\text{F1} \downarrow, \text{F2} = \bot, \text{F4} \uparrow, \text{F5} = \bot$
\item $\textbf{Action 2}$\\
      $\textit{pre: }\text{F1} = 0, \text{F2} = \bot, \text{F3} > 0, \text{F4} > 0, \text{F5} = \bot, \text{F6} = \top$\\
      $\textit{eff: }\text{F2} = \top, \text{F3} \downarrow$
\item $\textbf{Action 3}$\\
      $\textit{pre: }\text{F1} = 0, \text{F2} = \top, \text{F4} > 0, \text{F5} = \bot, \text{F6} = \top$\\
      $\textit{eff: }\text{F6} = \bot$
\item $\textbf{Action 4}$\\
      $\textit{pre: }\text{F1} =0, \text{F2} = \top, \text{F4} > 0, \text{F5} = \bot, \text{F6} = \bot$\\
      $\textit{eff: }\text{F2} = \bot$
\item $\textbf{Action 5}$\\
      $\textit{pre: }\text{F2} = \bot, \text{F4} > 0, \text{F5} = \bot, \text{F6} = \bot$\\
      $\textit{eff: }\text{F1} \uparrow, \text{F2} = \top, \text{F4} \downarrow, \text{F5} = \top$
\item $\textbf{Action 6}$\\
      $\textit{pre: }\text{F1} > 0, \text{F2} = \top, \text{F5} = \top, \text{F6} = \bot$\\
      $\textit{eff: }\text{F2} = \bot, \text{F5} = \bot$
\item $\textbf{Action 7}$\\
      $\textit{pre: }\text{F1} > 0, \text{F2} = \bot, \text{F3} > 0, \text{F4} =0, \text{F5} = \bot, \text{F6} = \bot$\\
      $\textit{eff: }\text{F2} = \top$
\item $\textbf{Action 8}$\\
      $\textit{pre: }\text{F1}  > 0, \text{F2} = \top, \text{F3} > 0, \text{F4} = 0, \text{F5} = \bot, \text{F6} = \bot$\\
      $\textit{eff: }\text{F6} = \top$
\item $\textbf{Action 9}$\\
      $\textit{pre: }\text{F1} > 0, \text{F2} = \top, \text{F3} > 0, \text{F4} = 0, \text{F5} = \bot, \text{F6} = \top$\\
      $\textit{eff: }\text{F2} = \bot, \text{F3} \uparrow$
\end{itemize}

% \begin{theorem}
% For any feature $\phi$ under complexity limit $M$, If for any named variable of $\phi$ the set of predicates with arguments containing that named variable is a subset of one stream's certified predicates and disjoint with all other stream's certified predicates, then the feature is probabilistically correct.
% \end{theorem}

% \begin{proof}
% See appendix
% \end{proof}

\end{document}